\documentclass[letterpaper]{article} 
\usepackage{aaai25}  
\usepackage{times}  
\usepackage{helvet}  
\usepackage{courier}  
\usepackage[hyphens]{url}  
\usepackage{graphicx} 
\usepackage{natbib}  
\usepackage{caption} 
\usepackage{algorithm}
\usepackage{algorithmic}

\usepackage{newfloat}
\usepackage{listings}
\DeclareCaptionStyle{ruled}{labelfont=normalfont,labelsep=colon,strut=off} 
\floatstyle{ruled}
\newfloat{listing}{tb}{lst}{}
\floatname{listing}{Listing}

\usepackage{graphicx} 
\usepackage{soul}
\usepackage{booktabs}
\usepackage{tikz}
\usetikzlibrary{arrows, backgrounds, fit, positioning, shapes.geometric}
\usepackage{natbib}
\usepackage{svg}
\usepackage{verbatim}
\usepackage{xurl}
\title{SHARPIE: A Modular Framework for Reinforcement Learning and Human-AI Interaction Experiments}
\author {
    Hüseyin Ayd{\i}n\textsuperscript{\rm 1}\equalcontrib,
    Kevin Godin-Dubois\textsuperscript{\rm 2}\equalcontrib,
    Libio Goncalvez Braz\textsuperscript{\rm 1}\equalcontrib,
    Floris den Hengst\textsuperscript{\rm 2}\equalcontrib,
    Kim Baraka\textsuperscript{\rm 2},
    Mustafa Mert Çelikok\textsuperscript{\rm 3},
    Andreas Sauter\textsuperscript{\rm 2},
    Shihan Wang\textsuperscript{\rm 1},
    Frans A. Oliehoek\textsuperscript{\rm 3}
}
\affiliations {
    \textsuperscript{\rm 1}Utrecht University, 
    \textsuperscript{\rm 2}Vrije Universiteit Amsterdam,
    \textsuperscript{\rm 3}TU Delft\\
    h.aydin@uu.nl, k.j.m.godin-dubois@vu.nl, l.b.goncalvesbraz@uu.nl, f.den.hengst@vu.nl, k.baraka@vu.nl, m.m.celikok@tudelft.nl, a.sauter@vu.nl, s.wang2@uu.nl, f.a.oliehoek@tudelft.nl\\
}

\usepackage{bibentry}

\begin{document}

\maketitle

\begin{abstract}
Reinforcement learning (RL) offers a general approach for modeling and training AI agents, including human-AI interaction scenarios. In this paper, we propose SHARPIE (\textbf{S}hared \textbf{H}uman-\textbf{A}I \textbf{R}einforcement Learning \textbf{P}latform for \textbf{I}nteractive \textbf{E}xperiments) to address the need for a generic framework to support experiments with RL agents and humans. Its modular design consists of a versatile wrapper for RL environments and algorithm libraries, a participant-facing web interface, logging utilities, deployment on popular cloud and participant recruitment platforms. It empowers researchers to study a wide variety of research questions related to the interaction between humans and RL agents, including those related to interactive reward specification and learning, learning from human feedback, action delegation, preference elicitation, user-modeling, and human-AI teaming. The platform is based on a generic interface for human-RL interactions that aims to standardize the field of study on RL in human contexts.
\end{abstract}

%
\begin{links}
    \link{Code}{https://github.com/libgoncalv/SHARPIE}
\end{links}

\section{Introduction}\label{sec:introduction}

Reinforcement learning (RL) refers to a family of algorithms in which agents learn from the consequences of their own interactions with an environment to try to maximize the long-term reward obtained~\cite{sutton1998}. The available compute has seen an increase in the use of RL methods in a wide variety of problems, many of which involve humans in some sense~\cite{den2020reinforcement}. RL agents interact with humans in a wide variety of ways: in some problem settings, RL agents observe humans to achieve a common goal or act on their behalf~\cite{natarajan2010multi}; in other settings, RL agents communicate with humans to provide services or support their decision making~\cite{zhao2021efficient};  human feedback can also be used to evaluate the value of the agents' actions in other settings \cite{knox2009interactively, christiano2017deep}.

These different types of interaction highlight the need to incorporate humans in the training and evaluation of RL agents. However, their interaction patterns remain limited to basic and unidirectional. In contrast, hybrid human-AI intelligence, as envisioned by~\citet{akata2020research} emphasizes the need for rich, diverse, and dynamic interaction patterns between humans and AI agents in order to effectively address problems which humans nor agents cannot solve independently. Although there are software packages to study the interaction of RL agents and their environments, little attention has been paid to training and evaluation of RL agents in diverse and dynamic interaction patterns that include both humans and other agents.

In particular, popular RL libraries such as Gymnasium \cite{brockman2016openai,towers2024gymnasium}, PettingZoo \cite{terry2021pettingzoo}, JaxMARL \cite{rutherford2024jaxmarl}, MO-Gymnasium \cite{alegre2022mo} or HIPPO-Gym \cite{taylor2023improving}, provide a framework to train RL agents in benchmarks of particular formulations of the RL problem. These libraries also provide an interface to include a custom domain in the training. 
However, none of these libraries by themselves specifically tackles rich, diverse and dynamic interactions among humans and RL agents such as multi-modal communications for shared observations, delegated actions and common goals
that we believe to be required for hybrid intelligence. 

To address this research gap, we propose SHARPIE (\textbf{S}hared \textbf{H}uman-\textbf{A}I \textbf{R}einforcement-Learning \textbf{P}latform for \textbf{I}nteractive \textbf{E}xperiments) as a platform for studying how multiple humans and RL agents interact and collaborate effectively. This modular platform streamlines the design and execution of experiments with humans and interactive RL agents by:
\begin{itemize}
 \item providing a versatile wrapper for popular RL, multi-agent RL, and multi-objective RL environments and algorithms
 \item supporting configurable communication channels between the human and the RL agents with various modalities
 \item offering logging services, deployment utilities, and participant recruitment platforms integration.
\end{itemize}
\noindent The platform aims to empower researchers to address research questions on the interaction of RL agents and humans, including those related to e.g. interpretability, interactive reward specification and learning, learning from human feedback, action delegation, preference elicitation, user-modeling, and human-AI teaming.
As SHARPIE aims to facilitate human-agent collaboration research in the most generic way possible, it may also prove useful for simulating interactive robotic tasks prior to their physical implementation. Hence, we believe that SHARPIE will also lead to new research questions in the field of human-robot interaction \cite{goodrich2008human}.

Apart from its potential impact on other fields, the main contribution of SHARPIE is to introduce a standard for RL-based human-agent interactions in a multi-agent setting in the same way that the Gymnasium API has become this standard for fully simulated RL environments. In this paper, we first present some motivating use cases to sketch the scope and intended outcomes of this project with tangible use cases, including their relation with cognitive science. Next, we describe the SHARPIE framework design principles, the challenges encountered, and the solutions implemented to tackle these. We close with a discussion of future ambitions.

\section{Motivating Use Cases}\label{sec:motivating_examples}

\begin{table*}[tbp]
 \centering
 \renewcommand*{\arraystretch}{1.2}
 \begin{tabular}{cllllp{.29\textwidth}}
 \toprule
  &
  & Human role(s)
  & Agent role(s)
  & Environment
  & Interface
  \\
 \midrule

  1& Reward annotation
  & $1$ annotator
  & $1$ learner
  & AMaze (discrete)$^1$
  & Video output with timed feedback
  \\

  2& Exploration
  & $1$ task specifier
  & $n$ learners
  & AMaze (hybrid)$^2$
  & Maze layout \& policy selection
  \\

  3&Teaching
  & $n$ teachers
  & $n$ students
  & Simple Spread$^3$
  & Demonstrations
  \\
  
  4&Action delegation
  & $1$ controller
  & $1$ learner
  & Mountain Car$^4$
  & Environment interactions
  \\

  5&Task specification
  & $1$ task specifier
  & $1$ learner
  & MineCraft$^5$
  & Text communication channel 
  \\
  
  6&Human-AI Teaming
  & \multicolumn{2}{c}{$n$ collaborators}
  & Simple Tag$^3$
  & Environment interactions, intentions
  \\


  7&Utility elicitation
  & $1$ stakeholder
  & $1$ learner
  & Deep Sea Treasure$^6$
  & Example policies
  \\

  8&Shared decision support
  & $n$ stakeholders
  & $1$ advisor
  & Water-reservoir$^7$
  & Environment interactions
  \\
  \bottomrule
 \end{tabular}
 \caption[Use case examples]{
  Use case examples.
  References are:
  1- \citet{godindubois2024amaze};
  2- \citet{godindubois2025edhucat};
  3- \citet{lowe2017multi};
  4- \citet{moore1990efficient};
  5- \citet{johnson2016malmo};
  6- \citet{vamplew2011empirical};
  7- \citet{pianosi2013tree}.
  }
 \label{table:usecases}
\end{table*}

Table \ref{table:usecases} provides practical examples of target use cases that illustrate the wide range of problems that SHARPIE is designed to deal with. It details for each use case the roles taken on by human(s) and agent(s), an example environment and an envisioned interaction interface. These example use cases are not comprehensive and we will only implement some of these initially. We provide them here for illustrative purposes.

The first use case, entitled `reward annotation', corresponds to a setting in which a human annotator supplies the agent with a dense reward signal based on observations of the agent interacting with an environment. This reward signal complements a sparse reward signal in a maze navigation task, and models a setting in which a human expert guides the agent in completing its task successfully. It requires a visual representation of the environment and human reward feedback components in the User Interface (UI). 
The second, `exploration', use case targets the same maze solving task but suggests richer platform capabilities, as it requires the participant to be able to define desired maze layouts and select promising policies interactively~\cite{godindubois2025edhucat}. This use-case highlights the use of a person to specify the task the agent should solve, which can be found in myriad examples of humans interacting with RL agents~\cite{chang2024human}.

The third `teaching' use case targets a multi-agent setting in which a number of agents need to learn from human demonstrators the task of covering a number of landmarks in the environment. This task invites researchers to study how goals (i.e. target locations) of the other agents can be estimated in a straightforward unidirectional way~\cite{10.1145/3687272.3688298}. The `action delegation' use case deals with the problem of handing over control between human and learner, and suggests the need for humans and learners to effectively model the other agents' intended actions, goals and capabilities in a bidirectional setting.

The `task specification' use case targets interactively guiding the agent towards a high-level goal by specifying (sub)tasks for sparse reward tasks or open-ended settings. This use case illustrates the need for communication channels between agents and humans to specify goals in e.g. free text or a formal representation of the task at hand~\cite{icarte2022reward,den2022reinforcement}. It allows studying the ability of humans to effectively express their (sub)tasks, as well as studying approaches to agree on the semantics of task specifications. 
The `human-AI teaming' use case suggests the ability for human and RL agents to bidirectionally communicate their observations, intentions or goals~\cite{zhu2024survey}. Through effective and efficient interactions among humans and agents, we can form AI-human teams to coordinate learning behaviors, for example to solve classical predator-prey tasks~\cite{lowe2017multi}. 

Finally, we consider two multiobjective decision-making problems. The first, `utility elicitation' use case addresses the need for humans to select from a set of optimal policies in a multi-objective setting. The agent learns the attainable trade-offs between the gathering of treasures and the amount of time spent under water in a sea environment. It then presents their trade-offs as a Pareto front of polices. The human stakeholder can then select a solution policy that optimizes the particular trade-off that maximizes \emph{utility}.

The multiobjective setting is generalized to multiple stakeholders in the final use case, in which the agent controls the amount of water released by an energy-generating dam. The different stake-holders, i.e., the inhabitants of an upstream village, a downstream village, and the dam operators, need to agree on a policy that balances the objectives of meeting energy and water demand while minimizes downstream and upstream costs of flooding. The individual utilities for these stakeholders are mined separately and then used to compute a set of policies that for some trade-off to inform the stakeholders as part of a shared decision-making process using example trajectories.

We close this section with a brief discussion of how the proposed framework supports the study of RL, cognitive sciences, and their interconnection. Firstly, the ability to collect human demonstrations, reward annotations, and task descriptions (use cases 1,2,3 and 5) allows the study of human decision making, task success, and task definitions in humans. Furthermore, collaboration between RL agents and humans (use cases 4 and 6) allows the study of social learning, trust and reciprocity dynamics, and theory of mind in hybrid human-AI settings~\cite{jara2019theory,pires2024reciprocity}. The impact of RL agents on group decision-making processes can be studied through use cases 6, 7 and 8~\cite{seo2017reinforcement}, while the presence of communication channels (5) allows the study of language, communication, and grounding~\cite{zhang2024towards}. We believe that the number of possible (inter)connections between RL and cognitive science that can be studied with SHARPIE will continue to grow as suitable environments, algorithms and UI components are contributed to the SHARPIE framework.

\section{Related Work}\label{sec:related_work}
Several successful software packages and platforms for the development of RL algorithms and evaluation environments have been made over the past decades, emphasizing a different number of objectives, different numbers of agents, different kinds of tasks and different programming languages, etc. We do not intend to review all software packages in RL and here we focus only on the related work that particularly deals with RL platforms that explicitly target the study of RL in a context that involves humans.

HIPPO-GYM \cite{taylor2023improving}, provides a framework that involves human interactions in a RL setting. However, this framework is limited to single-agent tasks and provides only specific interactions in those tasks (i.e. humans teaching RL agents). \citet{knierim2024prosody} presents a framework in human teaching RL agents setting with audio communication. Their codebase also facilitates a Wizard-of-Oz approach \cite{greenWoZ1985}, where a participant can play the role of a teacher while a researcher plays the role of an artificial agent. Although an alternative modality is proposed in this study, the interactions are designed to be one-way, with human teachers guiding RL-agents.

Another line of work presents Reinforcement Learning from Human Feedback (RLHF) platforms \cite{christiano2017deep, anonymous2023unirlhf}. However, the interfaces for human interaction provided by these platforms are only capable of handling elicited feedback for RLHF.

There are several software platforms for conducting behavioral experiments \cite{peirce2007psychopy,de2015jspsych}. None of these offers any infrastructure for any type of artificially intelligent entities, whether learning or not.

Various platforms and software packages have been developed for the deployment of RL \cite{gauci2018horizon,albers2022addressing,zhu2024pearl}. However, these are not focused on controlled experiments on the interaction between RL agents and humans.

Previous studies in multi-agent learning such as MAgent \cite{zheng2018magent} and MOMAland \cite{felten2024momaland} are capable of handling multi-agent experiments. MAgent supports a large population of agents, ranging from hundreds to millions, whereas it is possible to train the agents for multiple objectives in MOMAland. However, these packages do not yet support human interaction in any way. SHARPIE specifically targets the gap of easily setting up experiments with (multiple) human participants and RL agents by interfacing with existing environment and algorithm libraries.


One of the most related work to our study is Interactive Gym \cite{chase2024interactivegym} which is a framework in early stage of development that enables human-AI interaction in multi-agent RL settings. This framework currently provides two different setups: single human with multiple agents in a browser based execution and any number of human and AI agents in a client-server based execution of experiments. It is difficult to assess the exact differences between this framework and ours, as both are under development. However, with its current structure, Interactive Gym seems to be specific to the Gymnasium environments where human(s) can take the control of RL agents in a game setup which can require more effort to deal with a diverse set of scenarios including various type of human-RL agent interactions. SHARPIE, on the other hand, aims to provide a more generic and extendable design that is able to efficiently facilitate such scenarios. 

\section{SHARPIE Framework}\label{sec:sharpie_framework}

\begin{figure}[tbp]
 \includegraphics[width=\columnwidth]{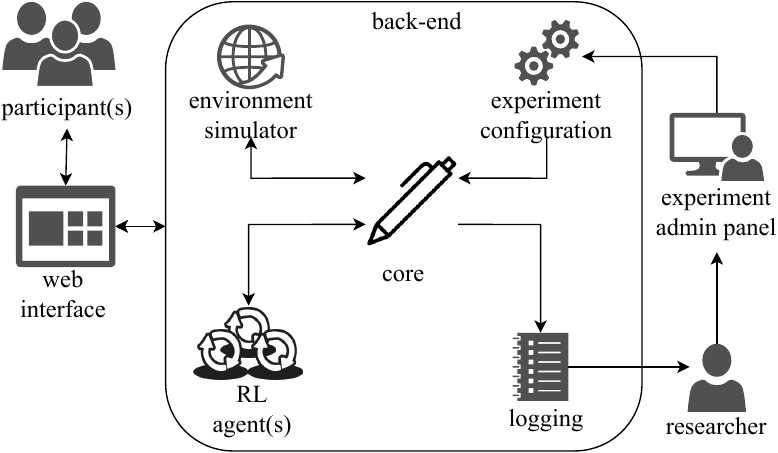}
 \caption{High-level SHARPIE architecture.}
 \label{fig:architecture}
\end{figure}

The SHARPIE library is a Python-based web framework that is currently under active development and that aims to provide a versatile wrapper around popular Reinforcement Learning a) environments, b) algorithms, and c) methodologies (see Figure \ref{fig:architecture}).
For the first part, it can encapsulate any existing environment that follows the conventional Gymnasium API (reset, step, render, etc.) which encompasses most of the existing RL platforms (see Appendix A).
Most importantly, the ambition of SHARPIE is not to be tightly integrated with any one particular environment or library, but rather to be compatible with many.

Furthermore, one of its core components is a customizable and easily deployable front-end UI which is primarily web-based and multi-modal.
This allows human users to interact with the environment and other (RL or human) agents.
Additionally, RL agents may be able to request action delegation, explicitly or as prompted by their learning algorithm, to further increase the range of interaction scenarios that SHARPIE can help streamline.

The front-end also provides various complementary utilities to further smoothen out the experimental processes: from (a)synchronous evaluations of a learning agent to scheduling and management on long-term data storage.
In the first case, this takes many forms, such as a preference elicitation module to visualize and rank trajectories, actions, and policies.
In the second case, this includes all the logging facilities required in a large-scale RL experiment, especially one involving human participants.
Indeed, it is of paramount importance to be able to securely and robustly store any relevant data, as restarting such an experiment from scratch might be, at best, long and costly and, at worse, practically impossible.

In addition, SHARPIE is designed to support multi-modal communication channels to allow communication between agents (again, RL or human).
Depending on said agents, these channels may be used for e.g. coordination, teaching, or overriding.
While initially designed for short semantic content, such as predefined sets of symbols, these communication mediums will be expanded upon to include any and all relevant formats (natural text, images, audio, etc.).

Finally, the library is designed with utilities to deploy to a cloud server, a private machine, or a local host.
Given the widespread support for Python, this encompasses any desktop operating system (e.g. Linux, Mac, Windows) with Python 3 installed, including remote web-servers or platforms such as Amazon Web Services (AWS). This provides an abstraction between the machines that the researchers and participants are using and where the experiment is actually running. Such a feature is essential in studies involving numerous participants with varying locations and hardware.


\section{Discussion \& Future Work}\label{sec:discussion_and_future_work}
We have presented and motivated the design and implementation of a framework to accelerate human–AI interaction studies, with a particular focus on studies for multi-agent tasks involving humans.
As stated, the scope of this project is not to implement alternative environment and algorithms for such studies, but rather to integrate with existing solutions seamlessly and provide utilities to ease experimentation with participants.
This is done by empowering researchers through a modular software interface relying, in part, on the de facto standard set by Gymnasium.
With SHARPIE we aim to provide an easily integrable framework that researchers can use to painlessly set up experiments involving both human and artificial agents.
Our hope is that in turn such an architecture lays the foundation for a standard for the interaction between human and artificial agents.

This modular approach also allows for numerous directions of improvement in terms of interoperability and features.
In the former case, we aim to provide an increasing number of ready-made, supported plugins to handle a large part of the existing work on environments, libraries, and deployment options.
In the latter case, we plan to widen the scope of possible human-agent interactions by incorporating additional modalities such as audio or video~\cite{christofi2024uncovering,knierim2024leveraging}.
The resulting real-time and multimodal communication between users and agents would allow the study of rich and fine-grained communication protocols, and support experiments involving pitch and tone, nonverbal cues, etc.
Finally, we envision a hosted version of SHARPIE that can be used for outreach, education and user literacy purposes.

\section*{Acknowledgements}
This research was funded by the Hybrid Intelligence Center, a 10-year programme funded by the Dutch Ministry of Education, Culture and Science through the Netherlands Organisation for Scientific Research, \url{https://hybridintelligence-centre.nl}, grant number 024.004.022.

\bibliography{references}

\onecolumn
\appendix
\section{Appendix A: Popular RL packages and their APIs}
\label{sec:popular_rl_packages}
\begin{table}[h!]
\centering
\begin{tabular}{llllll}
\toprule
Name & Description & Language & Target & API \\\midrule
Amaze$^1$ & Maze navigation environments & Python & Research & Gymnasium+ \\
CleanRL$^2$ & Algorithm library & Python/Torch & Research/Production & Gymnasium \\
Gymnasium$^3$ & De-facto standard environments & Python & Research & Gymnasium \\
HIPPO-Gym$^4$ & Human Input Parsing & Python & Research & Gymnasium+ \\
Interactive-Gym & Interactive web-based experiment platform & Python & Research & Gymnasium+ \\
JaxMARL$^5$ & Multi-agent environments & JAX & Research & Gymnasium+ \\
Momaland$^6$ & Multi-agent multi-objective environments & Python & Research & Gymnasium \\
MO-Gymnasium$^7$ & Multi-objective environments & Python & Research & Gymnasium+ \\
MORL Baselines$^8$ & Multi-objective algorithm library & Python & Research & Gymnasium+ \\
OpenSpiel$^9$ & Game (theory) environments/algorithms & Python & Research & Gymnasium \\
Pearl$^{10}$ & Algorithm library & Python & Production & Gymnasium \\
PettingZoo$^{11}$ & Multi-agent environments & Python & Research & Gymnasium+ \\
Ray RLlib$^{12}$ & Algorithm library & Python/Torch & Production & Gymnasium \\
RLax$^{13}$ & Algorithm library & JAX & Research & Gymnasium \\
Stable-baselines3$^{14}$ & Algorithm library & Python/Torch & Research/Production & Gymnasium \\
Uni-RLHF$^{15}$ & Annotation tool & Python & Annotation & Gymnasium+ \\
\bottomrule
\end{tabular}
\caption{Comparison of RL Packages, APIs, and Applications. $+$ indicates minor alterations to the Gymnasium API such as a vectorial reward for multi-objective, or vectorial action for multi-agent RL. References are: 1- \citet{godindubois2024amaze}; 2- \citet{huang2022cleanrl}; 3- \citet{brockman2016openai}; 4- \citet{taylor2023improving}; 5- \citet{rutherford2024jaxmarl}; 6-\citet{felten2024momaland}; 7- \citet{alegre2022mo}; 8- \citet{felten_toolkit_2023}; 9- \citet{LanctotEtAl2019OpenSpiel}; 10- \citet{zhu2024pearl}; 11- \citet{terry2021pettingzoo}; 12- \citet{pmlr-v80-liang18b}; 13- \citet{deepmind2020jax}; 14- \citet{stable-baselines3}; 15-\citet{anonymous2023unirlhf}.}
\label{tab:rl-packages}
\end{table}

\end{document}